%% file: paper.tex
\documentclass{article}
\usepackage{spconf,amsmath,graphicx}

\usepackage{lipsum}

\usepackage{rotating}
\input{newcommands2}

\input{newcommands3}

\usepackage{dblfloatfix}    
\usepackage{graphicx}
\usepackage{comment}
\usepackage{amsmath,amssymb} 
\usepackage{color}

\usepackage{slashbox}
\usepackage{times}
\usepackage{epsfig}
\usepackage{amssymb}
\usepackage{balance}

\usepackage{units}
\usepackage{mathtools}
\usepackage[linesnumbered,lined,boxed]{algorithm2e}
\usepackage{upgreek}
\usepackage{colortbl}
\usepackage{stmaryrd}
\usepackage[mathscr]{euscript}

\definecolor{morado}{cmyk}{0,1,0.50,0}
\definecolor{gris2}{cmyk}{0,0,0,0.25}
\definecolor{gris}{cmyk}{0,0,0,0.1}
\definecolor{amarillo}{cmyk}{0,0,0.6,0}
\definecolor{blanco}{cmyk}{0,0,0,0}
\definecolor{negro}{cmyk}{1,1,1,0}
\definecolor{orange}{cmyk}{0,0.5,0.8,0}

\newcommand{\blue}[1]{{\color{blue}{\textbf{#1}}}}
\newcommand{\red}[1]{{\color{red}{\textbf{#1}}}} 

\newcommand{\tscr}[1]{\boldsymbol{\mathscr{#1}}}
\newcommand{\nmat}[1]{\prescript{(n)}{}{\boldsymbol{#1}}} 

\usepackage{amsthm}
\usepackage{array}
\usepackage{multirow}
\usepackage{cite}

\theoremstyle{plain}
\newtheorem{theorem}{Theorem}[section]
\newtheorem{proposition}[theorem]{Proposition}

\theoremstyle{definition}

\theoremstyle{remark}

\title{Fast Unsupervised Tensor Restoration via Low-rank Deconvolution}
\name{David Reixach, Josep Ramon Morros\thanks{This work has been partially supported by the Spanish Ministry of Science and Innovation under the project DeeLight PID2020-117142GB-I00.}}
\address{Universitat Politècnica de Catalunya, BarcelonaTech, Spain}
\begin{document}
%
\maketitle
\begin{abstract}
Low-rank Deconvolution (LRD) has appeared as a new multi-dimensional representation model that enjoys important efficiency and flexibility properties. In this work we ask ourselves if this analytical model can compete against Deep Learning (DL) frameworks like Deep Image Prior (DIP) or Blind-Spot Networks (BSN) and other classical methods in the task of signal restoration. More specifically, we propose to extend LRD with differential regularization. This approach allows us to easily incorporate Total Variation (TV) and integral priors to the formulation leading to considerable performance tested on signal restoration tasks such image denoising and video enhancement, and at the same time benefiting from its small computational cost.
\end{abstract}
\begin{keywords}
Tensors, Restoration, Total Variation, Denoising, Enhancement
\end{keywords}
\section{Introduction}
\label{sec:intro}

The signal reconstruction problem generally relies on constraining the solution to be consistent with some prior knowledge. Traditional imaging priors include non-negativity, sparsity or self-similarity among others~\cite{rudin1992nonlinear,elad2006image,dabov2007image,heide2015fast,papyan2017convolutional}. More recently, the field has strongly been influenced by Deep Learning (DL), which offers a wide range of models to be considered as priors but mostly limited to imaging problems~\cite{ulyanov2018deep,lequyer2022fast, lee2022apbsn, li2023spatially,Pan_2023_ICCV}.


Recently, Low-rank Deconvolution (LRD)~\cite{reixach2023multi} has been introduced as a new model for compressed tensor representation and completion with promising results. However, as we show in this work, LRD lacks noise rejection abilities. To address this issue we propose to combine LRD together with a regularization function to solve tensor restoration problems. More specifically, by advantaging of the DFT formulation of LRD we propose to use this framework with implicit differential regularization in the DFT domain. This approach allows us to easily incorporate squared Total Variation (TV) and integral regularization into consideration by just solving a linear expression.


TV has been used as a prior in the past, alone~\cite{beck2009fast} or together with other priors such Deep Image Prior (DIP)~\cite{liu2019image}, both achieving state-of-the-art results in the task of image denoising. In this work we demonstrate how our combination of LRD and TV obtains remarkable performance on image denoising problems, benefiting also from being much lighter computationally than some DL and other classical denoising methods, and also extending to multi-dimensional data like tensors. As a by-product, we also show how this method can be used for video enhancement tasks. The main contributions of this work are as follows:

\begin{itemize}
    \item A novel method is proposed that extends the existing LRD framework with a new regularization function. It allows for tensor restoration tasks such denoising and detail enhancement in a fully unsupervised manner.
    \item A theoretical analysis is performed which allows incorporating the regularization function into the LRD framework by simply solving a linear expression in the DFT domain, allowing for efficient computation.
    \item The proposed method outperforms most of non-supervised state-of-the-art methods in both PSNR and execution time on image denoising. We also show that TV regularization based methods are still very competitive.
\end{itemize}

\section{Related work}
\label{sec:sota}

\subsection{Image Denoising}

Image denoising is a low-level computer vision topic that has been studied for decades and still remains open. It aims to obtain higher quality images from its corresponding noisy versions. Earlier approaches were based on modelling the signal or the noise and by formulating an optimization problem that tried to separate the two components. TV minimization~\cite{rudin1992nonlinear, beck2009fast} is a typical example of these approaches.

Following this idea, different solutions appeared that relied on signal decomposition in a sparse domain and filtering out the corresponding low-energy components~\cite{elad2006image}. Then the concept of image Non-local Self Similarity (NSS) appeared and boosted significantly image denoising performance, a prior assumption that is used by BM3D~\cite{dabov2007image}, EPLL~\cite{zoran2011learning} and WNNM~\cite{gu2014weighted} for example. A common thing that is shared by these classical methods in general is the fact that they are fully unsupervised, \ie, they only require a single noisy image to obtain its corresponding clean pair. Even so, these methods are computationally expensive, as NSS involves fragmenting the input image into different patches and performing exhaustive analysis between them.

During the last decade, many approaches based on DL have emerged. Initially, the vast majority of the methods~\cite{zhang2017learning} were based on fully supervised processes, \ie, requiring paired clean-noise image for training. However, to obtain such datasets that cover the whole range of noise possibilities is impractical. To get rid of this dependency, unsupervised methods based on DL priors such DIP~\cite{ulyanov2018deep} appeared. This method is able to learn natural image priors from large-scale datasets and requires only a single noisy image to obtain its corresponding clean pair. Similarly, Noise2Fast (N2F)~\cite{lequyer2022fast} trains on a discrete set of clean images and does not require any prior knowledge of the noise distribution. However, these methods rely on solving a DL based optimization problem for each sample, which makes them computationally expensive too.

More recently, a new category of methods has arisen, namely DL self-supervised. These methods do not require paired clean-noisy images, as they are able to learn from strictly noisy samples. Many of these approaches rely on a Blind-Spot Network (BSN) although in general they can only be applied under the assumption of pixel-wise independent noise which is unrealistic for a real world scenario. To tackle this issue, AP-BSN~\cite{lee2022apbsn} propose asymmetric Pixel-shuffle Downsampling (PD) factors and post-reﬁnement processing to make a better trade-off between noise removal and aliasing artifacts. SDAP~\cite{Pan_2023_ICCV} propose a new BSN framework in combination with a Random Sub-samples Generation (RSG) strategy that tries to break the pixel-wise independent noise assumption. Similarly, SASS~\cite{li2023spatially} propose a new strategy to break such assumption by taking into account the respective characteristics of ﬂat and textured regions in noisy images, and construct supervisions for them separately. However, self-supervised methods usually still require a large number of noisy images and in general they do not report good results when trained with scarce data.

\subsection{Detail Enhancement}

The goal of detail enhancement is to improve the visual appearance of images by increasing local contrast or amplifying details. Most existing techniques achieve it by modifying a decomposed version of the image. Such decomposition is typically achieved by filtering or optimization.

To that end, TV has been used in the past with great results~\cite{papyan2017convolutional}. Their idea is to decompose the image between two components: texture and cartoon. Texture is modelled as the signal which incorporates the main TV component, while cartoon holds the rest. Following this concept, our method, which is capable of such decomposition, eases the task of video enhancement (detail enhancement applied to tensors).


\section{Low-rank Deconvolution with Differential Regularization}
\subsection{Notation}
\label{sec:note}

Let $\tscr{J} \in \mathbb{R}^{I_1\times I_2 \times\ldots\times I_N}$ be a $N$-order tensor. The PARAFAC~\cite{harshman1970foundations} decomposition (a.k.a. CANDECOMP~\cite{carroll1970analysis}) is defined as:
\begin{equation}\label{eq:parafac}
	\tscr{J} \approx \sum_{r=1}^{R}\mu_r  \bv_r^{(1)} \circ \bv_r^{(2)} \circ \ldots \circ \bv_r^{(N)},
\end{equation}
where $\bv_r^{(n)} \in \mathbb{R}^{I_n} $ with $n=\{1,\ldots,N\}$ and $\mu_r \in \mathbb{R}$ with $r=\{1,\ldots,R\}$, represent an one-order tensor and a weight coefficient, respectively. $\circ$ denotes an outer product of vectors. Basically, Eq.~\eqref{eq:parafac} is a rank-$R$ decomposition of $\tscr{J}$ by means of a sum of $R$ rank-$1$ tensors. If we group all these vectors per mode $(n)$, as $\bX^{(n)} = \big[ \bv_1^{(n)}, \bv_2^{(n)}, \ldots, \bv_R^{(n)} \big]$, we can define the Kruskal operator~\cite{kolda2006multilinear} as follows: 
\begin{equation}\label{eq:kruskal}
\llbracket \bX^{(1)}, \bX^{(2)}, \ldots, \bX^{(N)} \rrbracket =  \sum_{r=1}^{R} \bv_r^{(1)} \circ \bv_r^{(2)} \circ \ldots \circ \bv_r^{(N)},
\end{equation}
being the same expression as Eq.~\eqref{eq:parafac} with $\mu_r=1$ for $\forall r$, i.e., depicting a rank-$R$ decomposable tensor.

For later computations, we also define a matricization transformation to express tensors in a matrix form. Particularly, we will use a special case of matricization known as $n$-mode matricization~\cite{bader2006algorithm,kolda2006multilinear}. To this end, let $\mathcal{C} = \{c_1,\ldots,c_G\} =\{1,\dots,n-1,n+1,\dots,N\}$ be the collection of ordered modes different than $n$, and $\Lambda = \nicefrac{\prod_t I_t}{I_n}$ be the product of its correspondent dimensions; we can express then a tensor $\tscr{K}$ in a matricized array as $\prescript{(n)}{}{\bK} \in \mathbb{R}^{I_n\times\Lambda}$. Note that we represent the $n$-mode matricization by means of a left super-index. The $n$-mode matricization is a mapping from the indices of $\tscr{K}$ to those of $\prescript{(n)}{}{\bK}$, defined as:
\begin{equation}\label{tensorm1}
	\big(\prescript{(n)}{}{\bK}\big)_{i_n,j} = \tscr{K}_{i_1,i_2,\dots,i_N} \, ,
\end{equation}
with:
\begin{equation}\label{tensorm2}
	j = 1+\sum_{g=1}^{G}\big[(i_{c_g}-1)\prod_{g'=1}^{G-1}I_{c_{g'}} \big].
\end{equation}

With these ingredients, and defining $\tscr{K} = \llbracket \bX^{(1)}, \dots, \bX^{(N)} \rrbracket$, by following~\cite{kolda2006multilinear} we can obtain the $n$-mode matricization of the Kruskal operator as: 
\begin{equation}\label{eq:kruskal_m}
\prescript{(n)}{}{\bK} = \bX^{(n)}(\bQ^{(n)})^\top, 
\end{equation}
with:
\begin{equation}\label{eq:Q}
\bQ^{(n)} = \bX^{(N)}\odot \ldots \odot  \bX^{(n+1)}\odot  \bX^{(n-1)}\odot \ldots \odot   \bX^{(1)},
\end{equation}
where $\odot$ denotes the Khatri-Rao product.

Finally, we can express the vectorized version of Eq.~\eqref{eq:kruskal_m} as: 
\begin{equation}
\label{eq:kruskal_v}
\mathrm{vec}\left(\prescript{(n)}{}{\bK}\right) =
\big[\bQ^{(n)}\otimes \bI_{I_n}\big]\mathrm{vec}(\bX^{(n)}),
\end{equation}
where $\otimes$ indicates the Kronecker product, and $\mathrm{vec}(\cdot)$ is a vectorization operator. It is worth noting that the vectorized form of the Kruskal operator is represented by a linear expression.

\subsection{Revisiting LRD}
\label{sec:lrd}

We recall the formulation of LRD~\cite{reixach2023multi}. Let $\tscr{S} \in \mathbb{R}^{I_1\times I_2\times\cdots\times I_N}$ be a multi-dimensional signal. Our goal is to obtain a multi-dimensional convolutional representation $\tscr{S}\approx\sum_m \tscr{D}_m \ast \tscr{K}_m$, where  $\tscr{D}_m \in \mathbb{R}^{L_1\times L_2\times\cdots\times L_N}$ acts as a dictionary, and $\tscr{K}_m \in \mathbb{R}^{I_1\times I_2\times\cdots\times I_N}$, the activation map, is a low-rank factored tensor (\ie a Kruskal tensor). If we write $ \tscr{K}_m =  \llbracket \bX_m^{(1)}, \bX_m^{(2)}, \ldots, \bX_m^{(N)} \rrbracket $ with $  \mathbf{X}_m^{(n)} \in \mathbb{R}^{I_n\times R} $, we can obtain a non-convex problem as:
\begin{align}\label{eq:main}
	\argmin_{\{\bX_m^{(n)}\}}
	\frac{1}{2}&\left\lVert \sum_{m = 1}^{M}	\tscr{D}_m \ast \llbracket \bX_m^{(1)},\ldots,\bX_m^{(N)} \rrbracket -\tscr{S} \right\rVert_2^2 \nonumber\\
	&+\Phi(\{\hat{\bX}_m^{(n)}\}),
\end{align}
where $\ast$ indicates a $N$-dimensional convolution and $\Phi(\{\hat{\bX}_m^{(n)}\})$ is a regularization term. According to~\cite{reixach2023multi} we propose to solve the problem for each Kruskal factor $(n)$ alternately. Algorithm~\ref{algorithm_MAIN} provides with more details.

Following~\cite{reixach2023multi}, we solve the LRD optimization problem in a DFT domain assuming that boundary effects are neglegible (\ie~relying on the use of filters of small spatial support). To this end, we denote by $\hat{\bA}$ an arbitrary variable $\bA$ in the DFT domain. Let $\hat{{\bD}}^{(n)}_m =\mathrm{diag}\big(\mathrm{vec}\big(\nmat{\hat{D}}_m\big)\big) \in \mathbb{R}^{\Lambda I_n\times\Lambda I_n}$ be a linear operator for computing convolution, and $\hat{\bx}_m^{(n)} = \mathrm{vec}(\hat{\bX}_m^{(n)})\in \mathbb{R}^{RI_n}$ be the vectorized Kruskal factor. We define $ \hat{\mathbf{Q}}_m^{(n)} =  \hat{\mathbf{X}}_m^{(N)}\odot \cdots \odot  \hat{\mathbf{X}}_m^{(n+1)}\odot  \hat{\mathbf{X}}_m^{(n-1)}\odot \cdots \odot \hat{\mathbf{X}}_m^{(1)} \in \mathbb{R}^{\Lambda\times R} $, as it was done in Eq.~\eqref{eq:Q}, with $\Lambda$ defined in section~\ref{sec:note}. Then, by using  Eq.~\eqref{eq:kruskal_v} we define:
\begin{align}
	\hat{\bW}_m^{(n)} &= \hat{\bD}_m^{(n)}\big[\hat{\bQ}_m^{(n)}\otimes \bI_{I_n}\big] \label{eq:dft_wm} ,\\
	\hat{\bW}^{(n)} &= \big[\hat{\bW}_0^{(n)},\hat{\bW}_1^{(n)},\ldots,\hat{\bW}_M^{(n)}\big] \label{eq:dft_w} ,\\
	\hat{\bx}^{(n)} &= \big[(\hat{\bx}_0^{(n)})^\top,(\hat{\bx}_1^{(n)})^\top,\ldots,(\hat{\bx}_M^{(n)})^\top\big]^\top \label{eq:dft_x},
\end{align}
All these algebraic modifications together with $\hat{\mathbf{s}}^{(n)} = \mathrm{vec}(\prescript{(n)}{}{\hat{\bS}})$  allow us to transform the problem~(\ref{eq:main}) into~(\ref{eq:admm_dft2}). And the solution will depend on the regularizer chosen ($\Phi(\{\hat{\bx}_m^{(n)}\})$):
\begin{equation}\label{eq:admm_dft2}
	\hspace{-0.2cm}\argmin_{\hat{\bx}^{(n)}}
	\frac{1}{2}\left\lVert	\hat{\bW}^{(n)} \hat{\bx}^{(n)} -\hat{\bs}^{(n)} \right\rVert_2^2
	+\Phi(\{\hat{\bx}_m^{(n)}\}).
\end{equation}

\IncMargin{1em}
\begin{algorithm}[t!]
\SetKwInOut{Input}{input}\SetKwInOut{Output}{output}
\Input{$\tscr{S}$, $\{\tscr{D}_m\}_{m=1}^M,\{\bX_{0,m}^{(n)}\}_{n=1,m=1}^{N,M}, R>0$ }
\Output{ $\{\bX_m^{(n)}\}_{n=1,m=1}^{N,M}$ }

\tcc{Initialize Kruskal Factors}
$\{\bX_m^{(n)}\} = \{\bX_{0,m}^{(n)}\} $\\  
\tcc{Main Loop, Eq.~\eqref{eq:main}}
\While{not converged}{
\For{$n = 1,\dots,N$}{
{\scriptsize $\bX_m^{(n)}=\argmin
    \frac{1}{2}\left\lVert \sum_{m = 1}^{M}	\tscr{D}_m \ast \llbracket \bX_m^{(1)},\ldots,\bX_m^{(N)} \rrbracket -\tscr{S} \right\rVert_2^2  +
    \Phi(\{\bX_m^{(n)}\})$}
}
}
\caption{ \textbf{LRD algorithm} solves the LRD problem by means of an alternated approach for every $n$-mode.}\label{algorithm_MAIN}
\end{algorithm}\DecMargin{1em}

\subsection{Differential Regularization in the DFT Domain}\label{section_for_DFT_domain}

Following the approach by~\cite{rudin1992nonlinear}, a squared Total Variation and an integral regularizer can be added to the global problem in the following manner:
\begin{align}\label{eq:lrd_tv}
    \argmin_{\{\bX_m^{(n)}\}, \tscr{U}}
    \frac{1}{2}&\left\lVert \tscr{U} -\tscr{S} \right\rVert_2^2\nonumber + \frac{\gamma}{2}\left\lVert \tscr{U} \right\rVert_{TV}^2 + \frac{\zeta}{2}\left\lVert \tscr{U} \right\rVert_{TI}^2 + \Psi(\{\bX_m^{(n)}\})\nonumber .\\ 
    &\hspace{-1cm}\textrm{subject to} \hspace{0.5cm} 
    \tscr{U} = \sum_{m = 1}^{M}	\tscr{D}_m \ast \llbracket \bX_m^{(1)},\ldots,\bX_m^{(N)} \rrbracket
\end{align}
Where $\gamma$, $\zeta$ are parameters, and $\Psi(\{\bX_m^{(n)}\})$ is a regularization term. Then, with a little algebraic manipulation, and making use of the derivative and integral properties of the DFT, we can present the following proposition:
\begin{proposition}
\label{prop:tv}
The problem presented in eq.~(\ref{eq:lrd_tv}) with $\Psi(\{\bX_m^{(n)}\}) = \sum_{m = 1}^{M}\sum_{n = 1}^{N}\frac{\alpha}{2} \left\lVert \bX_m^{(n)} \right\rVert_2^2$ has a solution given by a linear expression in the DFT domain given by:
\begin{align}\label{eq:lrd_sol_tv}
	&\hspace{-0.2cm}\big[(\hat{\bW}^{(n)})^H \hat{\bW}^{(n)}+\gamma (\hat{\Theta}^{(n)})^H\hat{\Theta}^{(n)}\nonumber\\&+\zeta (\hat{\Omega}^{(n)})^H\hat{\Omega}^{(n)}+\alpha \bI_{\beta}\big]
	\hat{\bx}^{(n)}=(\hat{\bW}^{(n)})^H \hat{\bs}^{(n)},
\end{align}
where we have made use of $\hat{\mathbf{s}}^{(n)}$,  $\hat{\bx}^{(n)}$ and $\hat{\bW}^{(n)}$ defined in section~\ref{sec:lrd}. And defining:
\begin{align}
    (\hat{\Theta}_i^{(n)})^T = 2\pi j\xi_i \oplus\hat{\bW}^{(n)}\label{eq:theta_i}\\
    (\hat{\Omega}_i^{(n)})^T = (2\pi j\xi_i)^{-1} \oplus\hat{\bW}^{(n)},\\
    \hat{\Theta} = \big[\hat{\Theta}_0^{(n)}, \hat{\Theta}_1^{(n)}, \ldots, \hat{\Theta}_N^{(n)} \big],\\
    \hat{\Omega} = \big[\hat{\Omega}_0^{(n)}, \hat{\Omega}_1^{(n)}, \ldots, \hat{\Omega}_N^{(n)} \big]\label{eq:omega}
\end{align}
with $\xi_i$ being the vector of frequencies for the $i$-dimension, and $\oplus$ denoting element-wise product.
The problem is equivalent to the problem of eq.~(\ref{eq:main}) with:
\begin{align}
\Phi(\{\bx_m^{(n)}\}) &= \frac{\gamma}{2}\left\lVert (\hat{\Theta}^{(n)})^T\hat{\bx}^{(n)}\right \rVert_{2}^2\nonumber\\
&\hspace{2pt}+ \frac{\zeta}{2}\left\lVert (\hat{\Omega}^{(n)})^T\hat{\bx}^{(n)}\right \rVert_{2}^2 + \Psi(\{\bX_m^{(n)}\}).
\end{align}

\end{proposition}

\begin{proof} 
The squared total variation regularization is given by
\begin{equation}
    \frac{\gamma}{2}\left\lVert \tscr{U} \right\rVert_{TV}^2 = \frac{\gamma}{2}\left\lVert \nabla\tscr{U} \right\rVert_{2}^2,
\end{equation}
where, 
\begin{align}
    &\frac{\gamma}{2}\left\lVert \nabla\tscr{U} \right\rVert_{2}^2 =\\     
    &\frac{\gamma}{2}\left\lVert \left[\left(\frac{\partial\bu^{(n)}}{\partial \bt_0}\right)^T, \left(\frac{\partial\bu^{(n)}}{\partial \bt_1}\right)^T, \dots, \left(\frac{\partial\bu^{(n)}}{\partial \bt_N}\right)^T\right]^T\right\rVert_{2}^2.\nonumber
\end{align}
By using the derivative property of the DFT transform, 
\begin{align}
    \mathcal{F}\biggl\{\frac{\partial\bu^{(n)}}{\partial \bt_i}\biggr\}&= 2\pi j\xi_i\oplus\mathcal{F}\{\bu^{(n)}\} \nonumber\\
     &=2\pi j\xi_i\oplus\hat{\bW}^{(n)} \hat{\bx}^{(n)},
\end{align}
with $\xi_i$ and $\oplus$ defined in section~\ref{sec:lrd}. Together with the definition of $\hat{\Theta}^{(n)}$ leads us to the following expression:
\begin{align}
    \mathcal{F}\Biggr\{\frac{\gamma}{2}&\left\lVert \sum_{m = 1}^{M}	\tscr{D}_m \ast \llbracket \bX_m^{(1)},\ldots,\bX_m^{(N)} \rrbracket\right\rVert_{TV}^2\Biggl\}= \nonumber\\
    &\frac{\gamma}{2}\left\lVert (\hat{\Theta}^{(n)})^T\hat{\bx}^{(n)}\right \rVert_{2}^2,
\end{align}
which can be derived in the following manner (complex derivative),
\begin{equation}\label{eq:proof_d}
    \frac{\partial\frac{\gamma}{2}\left\lVert (\hat{\Theta}^{(n)})^T\hat{\bx}^{(n)}\right \rVert_{2}^2}{\partial(\hat{\bx}^{(n)})^H} = \frac{\gamma}{2} (\hat{\Theta}^{(n)})^H\hat{\Theta}^{(n)}.
\end{equation}
The result above gives the solution for the derivative component. The proof for the squared integral component follows in the same manner and is given by:
\begin{align}
    &\frac{\zeta}{2}\left\lVert \tscr{U} \right\rVert_{TI}^2 =\\     
    &\frac{\zeta}{2}\left\lVert \left[\left(\int\bu^{(n)}\,d\bt_0\right)^T, \dots, \left(\int\bu^{(n)}\,d\bt_N\right)^T\right]^T\right\rVert_{2}^2.\nonumber
\end{align}
By using the integration property of the DFT transform, 
\begin{align}
    \mathcal{F}\biggl\{\int\bu^{(n)}\,d\bt_i\biggr\}&= (2\pi j\xi_i)^{-1}\oplus\mathcal{F}\{\bu^{(n)}\} \nonumber\\
     &=(2\pi j\xi_i)^{-1}\oplus\hat{\bW}^{(n)} \hat{\bx}^{(n)},
\end{align}
with $\xi_i$ and $\oplus$ defined in section~\ref{sec:lrd}. Together with the definition of $\hat{\Omega}^{(n)}$ leads us to the following expression:
\begin{align}
    \mathcal{F}\Biggr\{\frac{\zeta}{2}&\left\lVert \sum_{m = 1}^{M}	\tscr{D}_m \ast \llbracket \bX_m^{(1)},\ldots,\bX_m^{(N)} \rrbracket\right\rVert_{TI}^2\Biggl\}= \nonumber\\
    &\frac{\zeta}{2}\left\lVert (\hat{\Omega}^{(n)})^T\hat{\bx}^{(n)}\right \rVert_{2}^2,
\end{align}
which can be derived in the following manner (complex derivative),
\begin{equation}
    \frac{\partial\frac{\zeta}{2}\left\lVert (\hat{\Omega}^{(n)})^T\hat{\bx}^{(n)}\right \rVert_{2}^2}{\partial(\hat{\bx}^{(n)})^H} = \frac{\zeta}{2} (\hat{\Omega}^{(n)})^H\hat{\Omega}^{(n)}.
\end{equation}
The result above together with eq.~(\ref{eq:proof_d}) combined with the solution of eq.~(\ref{eq:main}), brings us to eq.~(\ref{eq:lrd_sol_tv}).

\end{proof}

\subsection{Image Denoising \& Detail Enhancement}\label{section:application}
The application to image denoising as a TV regularizer is straight-forward as it only involves solving~(\ref{eq:lrd_tv}) with a proper selection for $\gamma$ and $\alpha$, while $\zeta$ is left to $0.0$. Regarding the application to detail enhancement, let us reformulate~(\ref{eq:lrd_tv}) into the following expression:
\begin{align}\label{eq:lrd_tv_m}
    \argmin_{\{\bX_m^{(n)}\}, \{\tscr{U}_m\}}
    \frac{1}{2}&\left\lVert \sum_{m=1}^M\tscr{U}_m -\tscr{S} \right\rVert_2^2\nonumber + \sum_{m=1}^M\frac{\gamma_m}{2}\left\lVert \tscr{U}_m \right\rVert_{TV}^2\nonumber\\ &+ \sum_{m=1}^M\frac{\zeta_m}{2}\left\lVert \tscr{U}_m \right\rVert_{TI}^2 + \Psi(\{\bX_m^{(n)}\})\nonumber ,\\ 
    &\hspace{-1.8cm}\textrm{subject to} \hspace{0.5cm} 
    \tscr{U}_m = \tscr{D}_m \ast \llbracket \bX_m^{(1)},\ldots,\bX_m^{(N)} \rrbracket
\end{align}
which correspond to the same equation but with explicit signal $\tscr{U}_m$ reconstruction for each filter $m$. The idea of enhancement is to solve~(\ref{eq:lrd_tv_m}) and reconstruct the enhanced signal as $\tilde{\tscr{U}} = \sum_{m=1}^M\delta_m\tscr{D}_m \ast \llbracket \bX_m^{(1)},\ldots,\bX_m^{(N)}\rrbracket+\tscr{S}$ with a proper set of parameters $\{\gamma_m\}_{m=1}^M$, $\{\zeta_m\}_{m=1}^M$, $\{\delta_m\}_{m=1}^M$ and $\alpha$.

\begin{table*}[t!]
\centering
 \resizebox{17.4cm}{!} {
\begin{tabular}{ c | c || c | c | c | c | c | c | c | c | c | c | c | c ||}
\cline{2-14}
\multirow{19}{*}{
\resizebox{1.69cm}{!} {\includegraphics[viewport=690 25 750 780,clip, angle=0]{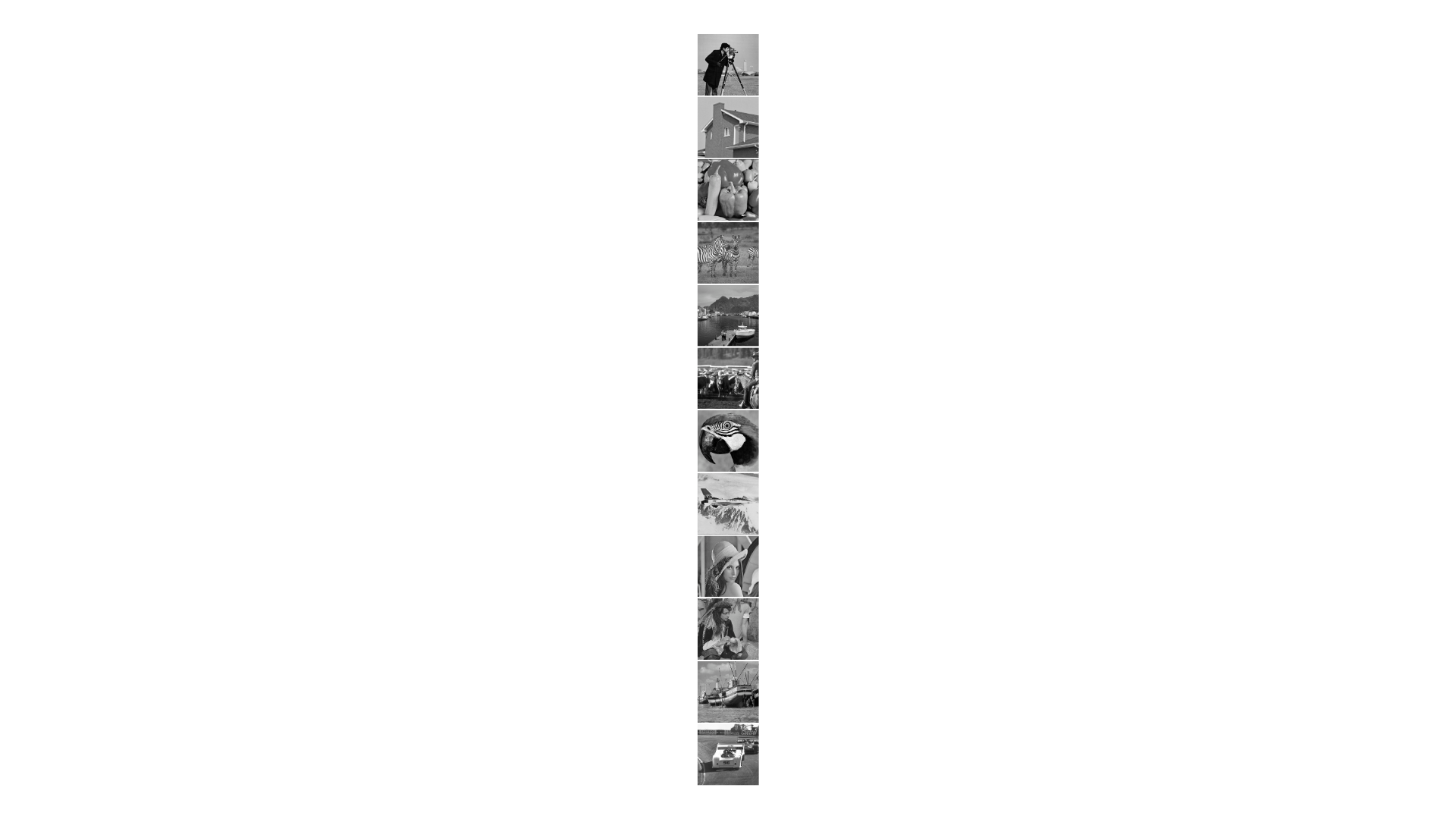}}}&{Images} & {1} & {2} & {3} & {4} & {5} & {6} & {7} & {8} & {9} & {10} & {11} & {12}\\
\cline{2-14}\multicolumn{14}{c}{}\\[-2ex]\cline{2-14}
&\multicolumn{13}{c||}{Input PSNR = $15.36$ dB / $\sigma=30$}\\
\cline{2-14}
&{BM3D} & 17.96 & 22.59 & 24.68 & 18.64 & 21.32 & 22.86 & 22.59 & 18.58 & 22.76 & 22.48 & 23.14 & 19.37\\
\cline{2-14}
&{EPLL} & 20.76 & 24.36 & 24.67 & \red{21.80} & 24.13 & 23.84 & 22.38 & 20.98 & 24.34 & 22.72 & 24.33 & 20.52\\
\cline{2-14}
&{TV} & 21.33 & \red{27.81} & \red{26.47} & 19.99 & 23.58 & 25.49 & \red{24.48} & \red{21.39} & \blue{26.66} & \blue{25.34} & 24.10 & 19.96\\
\cline{2-14}
&{WNNM} & 21.52 & \blue{29.67} & \blue{28.34} & \blue{23.54} & \blue{26.12} & \blue{28.65} & \blue{26.44} & \blue{24.30} & \red{26.29} & \red{24.91} & \blue{24.73} & \blue{25.78}\\
\cline{2-14}
&{DIP} & \blue{25.40} & 24.78 & 23.53 & 18.91 & 19.16 & 20.30 & 20.08 & 13.46 & 18.15 & 18.11 & 19.09 & 16.66\\
\cline{2-14}
&{N2F} & 20.19 & 17.72 & 19.59 & 17.78 & 15.78 & 16.79 & 17.82 & 15.65 & 13.55 & 10.74 & 18.22 & 16.62\\
\cline{2-14}
&{AP-BSN} & 19.51 & 21.84 & 21.72 & 17.62 & 22.13 & 23.80 & 20.13 & 15.15 & 22.02 & 20.37 & 22.11 & 17.52\\
\cline{2-14}
&{SDAP} & 15.17 & 17.96 & 10.75 & 16.78 & 16.52 & 13.00 & 15.66 & 16.68 & 17.66 & 17.65 & 14.26 & 16.65\\
\cline{2-14}
&{SASS} & 17.82 & 15.99 & 18.66 & 15.76 & 18.00 & 19.60 & 17.46 & 11.04 & 17.80 & 16.98 & 18.10 & 13.74\\
\cline{2-14}
&{LRD} & 16.86 & 15.69 & 18.53 & 16.55 & 17.95 & 18.61 & 18.10 & 13.09 & 18.48 & 17.43 & 18.60 & 16.11\\
\cline{2-14}
&{LRD-TV (Ours)} & \red{21.56} & 25.57 & 23.00 & 20.56 & \red{24.69} & \red{25.68} & 23.74 & 20.33 & 25.80 & 24.73 & \red{24.42} & \red{22.13}\\
\cline{2-14}\multicolumn{14}{c}{}\\[-2ex]\cline{2-14}
&\multicolumn{13}{c||}{Input PSNR = $12.18$ dB / $\sigma=50$}\\
\cline{2-14}
&{BM3D} & 15.69 & 15.32 & 21.00 & 16.57 & 18.47 & 19.68 & 19.44 & 14.28 & 19.20 & 17.53 & 20.35 & 15.51\\
\cline{2-14}
&{EPLL} & 19.17 & 21.50 & 20.68 & 18.52 & 20.48 & 19.81 & 18.75 & 18.52 & 20.30 & 20.02 & 19.93 & 18.82\\
\cline{2-14}
&{TV} & \red{20.41} & 24.23 & 21.30 & 18.17 & \blue{23.41} & \red{23.80} & \blue{22.25} & \red{18.86} & \red{23.48} & \red{22.69} & \red{21.32} & 19.45\\
\cline{2-14}
&{WNNM} & \blue{21.45} & \blue{26.92} & \blue{22.09} & \blue{20.86} & 21.26 & \blue{25.63} & \red{21.99} & \blue{21.92} & 22.41 & 21.73 & 18.00 & \blue{24.45}\\
\cline{2-14}
&{DIP} & 12.62 & 20.82 & 19.56 & 17.38 & 20.88 & 22.10 & 20.13 & 15.97 & 19.72 & 19.27 & 19.35 & 16.79\\
\cline{2-14}
&{N2F} & 17.83 & 13.71 & 15.17 & 12.76 & 13.19 & 13.28 & 14.04 & 12.99 & 12.12 & 8.52 & 14.45 & 13.99\\
\cline{2-14}
&{AP-BSN} & 17.53 & 16.42 & 17.11 & 14.69 & 16.23 & 17.67 & 16.08 & 11.13 & 15.81 & 16.35 & 15.23 & 14.65\\
\cline{2-14}
&{SDAP} & 12.47 & 14.24 & 8.37 & 11.95 & 13.34 & 11.50 & 13.25 & 12.93 & 13.80 & 15.96 & 12.16 & 12.69\\
\cline{2-14}
&{SASS} & 15.78 & 13.63 & 14.83 & 13.49 & 14.19 & 15.46 & 14.01 & 8.95 & 13.83 & 14.55 & 13.08 & 12.39\\
\cline{2-14}
&{LRD} & 14.08 & 13.21 & 14.49 & 13.68 & 13.75 & 14.91 & 14.35 & 9.00 & 13.61 & 14.10 & 14.17 & 12.98\\
\cline{2-14}
&{LRD-TV (Ours)} & 19.60 & \red{24.35} & \red{22.07} & \red{19.16} & \red{23.32} & 23.75 & \red{21.99} & 17.87 & \blue{23.77} & \blue{22.73} & \blue{23.07} & \red{21.04}\\
\cline{2-14}\multicolumn{14}{c}{}\\[-2ex]\cline{2-14}
&\multicolumn{13}{c||}{Input PSNR = $10.49$ dB / $\sigma=70$}\\
\cline{2-14}
&{BM3D} & 14.84 & 14.64 & 18.11 & 14.50 & 16.21 & 16.63 & 15.96 & 11.63 & 17.73 & 16.73 & 17.53 & 14.00\\
\cline{2-14}
&{EPLL} & 17.32 & 19.37 & 17.92 & 16.82 & 18.97 & 18.46 & 16.38 & 16.90 & 18.04 & 17.30 & 18.62 & 17.65\\
\cline{2-14}
&{TV} & \blue{19.12} & \blue{24.87} & \red{20.16} & 17.88 & \blue{22.91} & \red{23.37} & \red{20.82} & \red{19.72} & \blue{22.86} & \red{21.66} & \red{20.43} & 19.69\\
\cline{2-14}
&{WNNM} & 15.02 & 21.82 & 15.51 & \blue{19.06} & 18.65 & \blue{23.50} & \blue{20.91} & \blue{20.78} & 19.11 & 19.46 & 16.49 & \blue{22.99}\\
\cline{2-14}
&{DIP} & 13.19 & 19.74 & 19.39 & 17.08 & 19.47 & 21.16 & 18.41 & 14.16 & 18.31 & 17.52 & 16.87 & 15.40\\
\cline{2-14}
&{N2F} & 16.47 & 12.13 & 14.03 & 11.11 & 12.28 & 10.90 & 12.05 & 10.76 & 10.96 & 7.35 & 11.76 & 10.88\\
\cline{2-14}
&{AP-BSN} & 16.03 & 12.78 & 13.18 & 13.17 & 13.71 & 15.87 & 12.90 & 8.85 & 12.30 & 11.82 & 12.90 & 12.83\\
\cline{2-14}
&{SDAP} & 10.36 & 13.37 & 6.94 & 10.58 & 11.15 & 10.50 & 10.13 & 11.39 & 11.16 & 14.44 & 11.22 & 10.41\\
\cline{2-14}
&{SASS} & 15.08 & 11.48 & 12.01 & 12.22 & 12.49 & 14.17 & 11.61 & 7.66 & 11.32 & 10.94 & 11.68 & 11.39\\
\cline{2-14}
&{LRD} & 12.53 & 11.02 & 12.91 & 11.92 & 12.25 & 12.79 & 12.13 & 7.96 & 11.81 & 11.62 & 12.01 & 11.00\\
\cline{2-14}
&{LRD-TV (Ours)} & \red{18.27} & \red{23.46} & \blue{20.98} & \red{18.42} & \red{22.28} & 23.13 & 20.58 & 17.43 & \red{22.75} & \blue{21.90} & \blue{21.99} & \red{19.84}\\
\cline{2-14}\multicolumn{14}{c}{}\\[-2ex]\cline{2-14}
&\multicolumn{13}{c||}{Input PSNR = $9.49$ dB / $\sigma=90$}\\
\cline{2-14}
&{BM3D} & 14.50 & 13.02 & 17.69 & 13.74 & 15.82 & 16.29 & 15.26 & 11.14 & 16.28 & 13.60 & 15.75 & 12.87\\
\cline{2-14}
&{EPLL} & 17.01 & 18.16 & 16.18 & 15.46 & 17.83 & 16.61 & 15.26 & 16.00 & 16.80 & 16.68 & 17.44 & 16.32\\
\cline{2-14}
&{TV} & \blue{18.67} & \blue{22.92} & \red{20.04} & 16.98 & \blue{22.62} & 18.77 & \blue{20.77} & 17.05 & \blue{21.85} & \blue{21.33} & \red{20.58} & 18.95\\
\cline{2-14}
&{WNNM} & 14.26 & 19.23 & 15.11 & \blue{17.96} & 18.16 & \blue{21.91} & \red{20.43} & \blue{19.53} & 19.06 & 17.74 & 15.68 & \blue{22.19}\\
\cline{2-14}
&{DIP} & 10.32 & 17.00 & 15.74 & 16.21 & 20.74 & 21.01 & 17.34 & 13.17 & 18.25 & 16.75 & 16.40 & 15.50\\
\cline{2-14}
&{N2F} & 15.95 & 10.76 & 12.69 & 10.03 & 10.76 & 9.74 & 11.12 & 9.81 & 9.52 & 6.19 & 10.05 & 10.33\\
\cline{2-14}
&{AP-BSN} & 15.76 & 11.02 & 10.64 & 11.45 & 11.76 & 13.50 & 11.61 & 7.23 & 10.45 & 11.05 & 10.74 & 10.68\\
\cline{2-14}
&{SDAP} & 9.21 & 11.76 & 5.80 & 9.19 & 10.28 & 9.09 & 9.65 & 10.10 & 9.34 & 14.08 & 10.02 & 9.12\\
\cline{2-14}
&{SASS} & 15.07 & 10.11 & 9.86 & 10.51 & 10.81 & 12.12 & 10.44 & 6.56 & 9.72 & 10.19 & 10.01 & 9.73\\
\cline{2-14}
&{LRD} & 12.55 & 9.90 & 10.95 & 10.88 & 10.60 & 11.97 & 10.70 & 7.03 & 10.11 & 10.09 & 9.92 & 9.96\\
\cline{2-14}
&{LRD-TV (Ours)} & \red{17.77} & \red{22.41} & \blue{20.57} & \red{17.75} & \red{21.66} & \red{21.45} & 19.79 & \red{17.17} & \red{21.58} & \red{21.17} & \blue{21.13} & \red{19.38}\\
\cline{2-14}

\end{tabular}}
\vspace{0.01cm}
\caption{\textbf{Quantitative evaluation on image denoising.} \textbf{Left:} From top to bottom, we display the collection of twelve images chosen for the experiment. \textbf{Right:} The table reports the PSNR in dB (higher is better) using ten state-of-the-art approaches and our method for the task of image denoising. Results are reported for different levels of input noise. Best viewed in color.}
\label{table_denoising}
\end{table*}

\begin{figure*}[t!]
\centering
\resizebox{17.4cm}{!} {
\begin{tabular}{@{}ccccc@{}}
 \includegraphics[viewport=50 180 560 620, clip, angle=0]{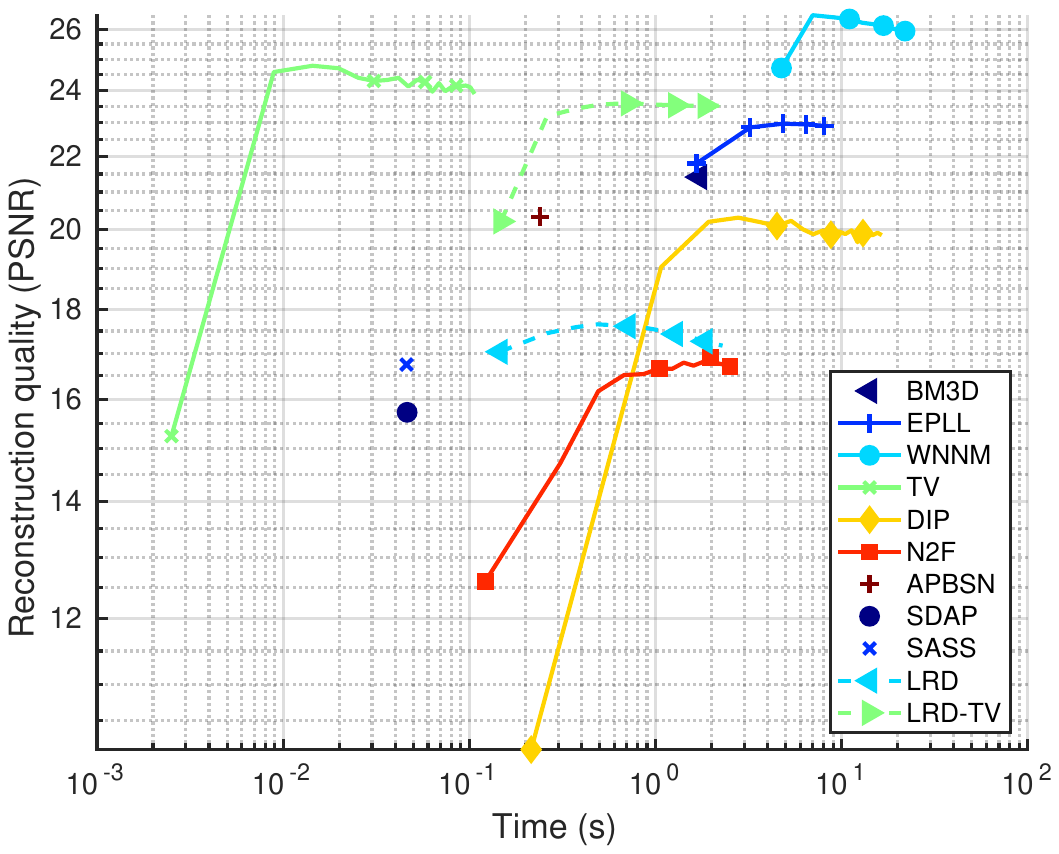}&
 \includegraphics[viewport=50 180 560 620, clip, angle=0]{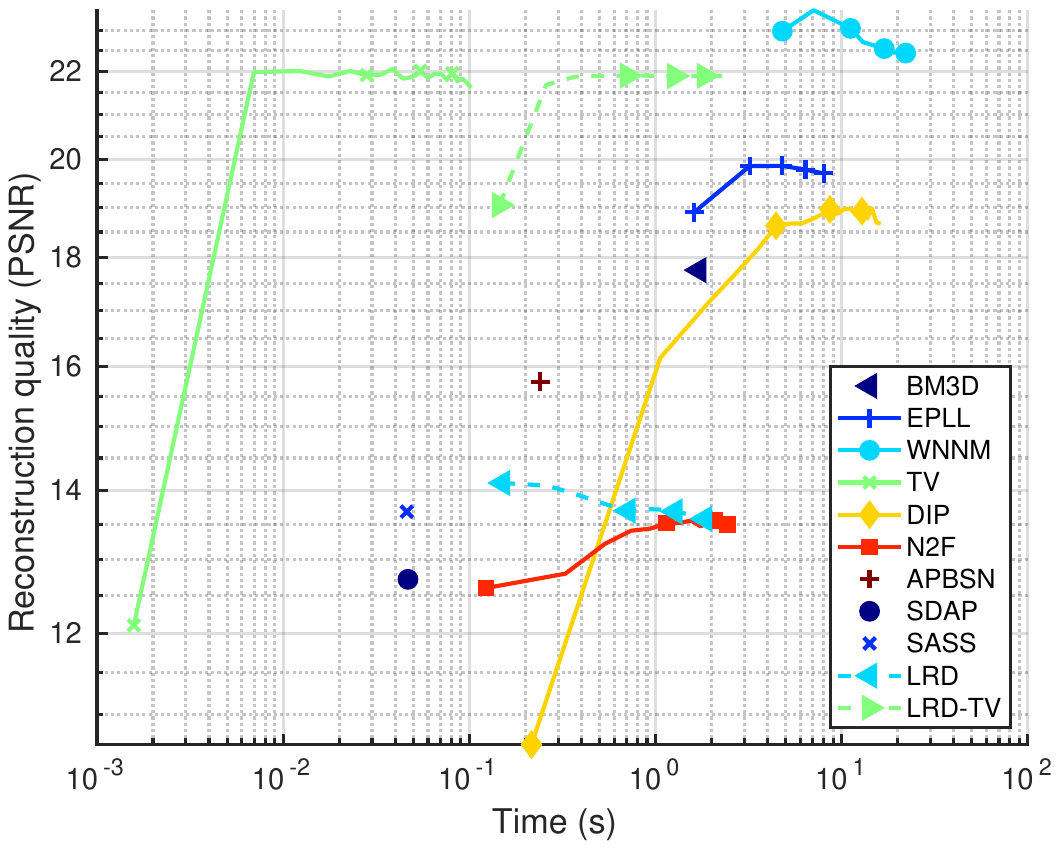}&
 \includegraphics[viewport=50 180 560 620, clip, angle=0]{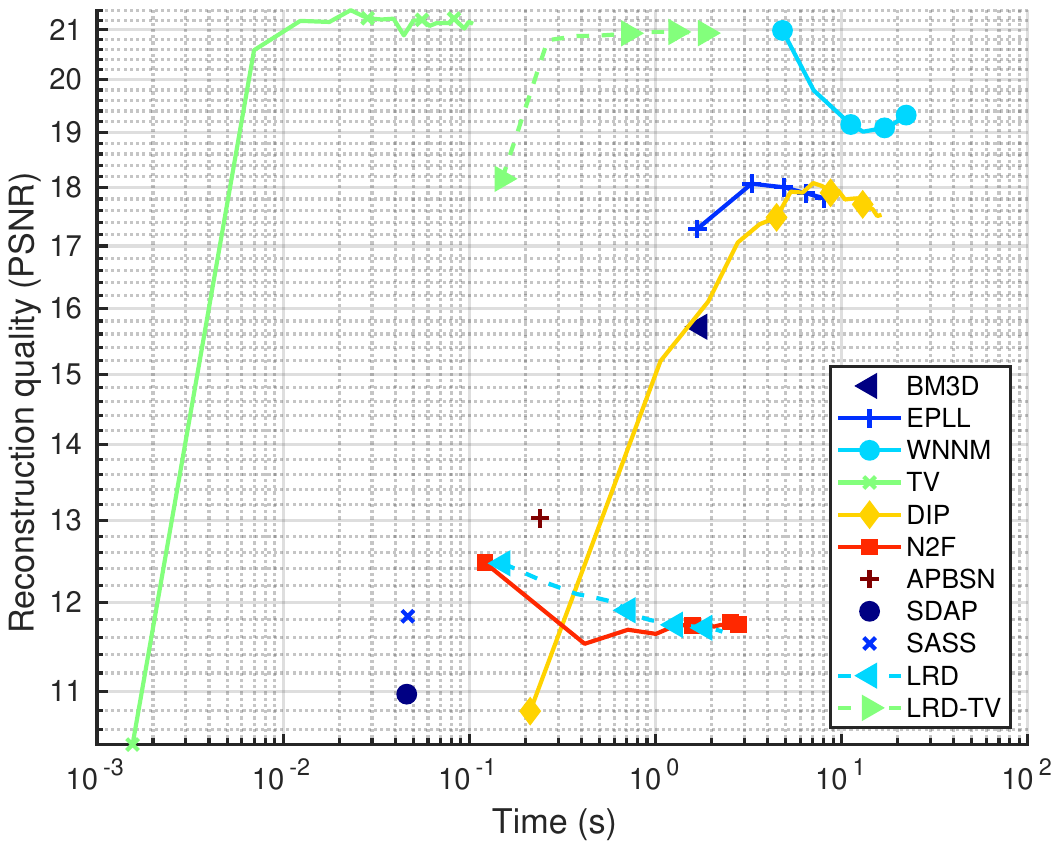}&
 \includegraphics[viewport=50 180 560 620, clip, angle=0]{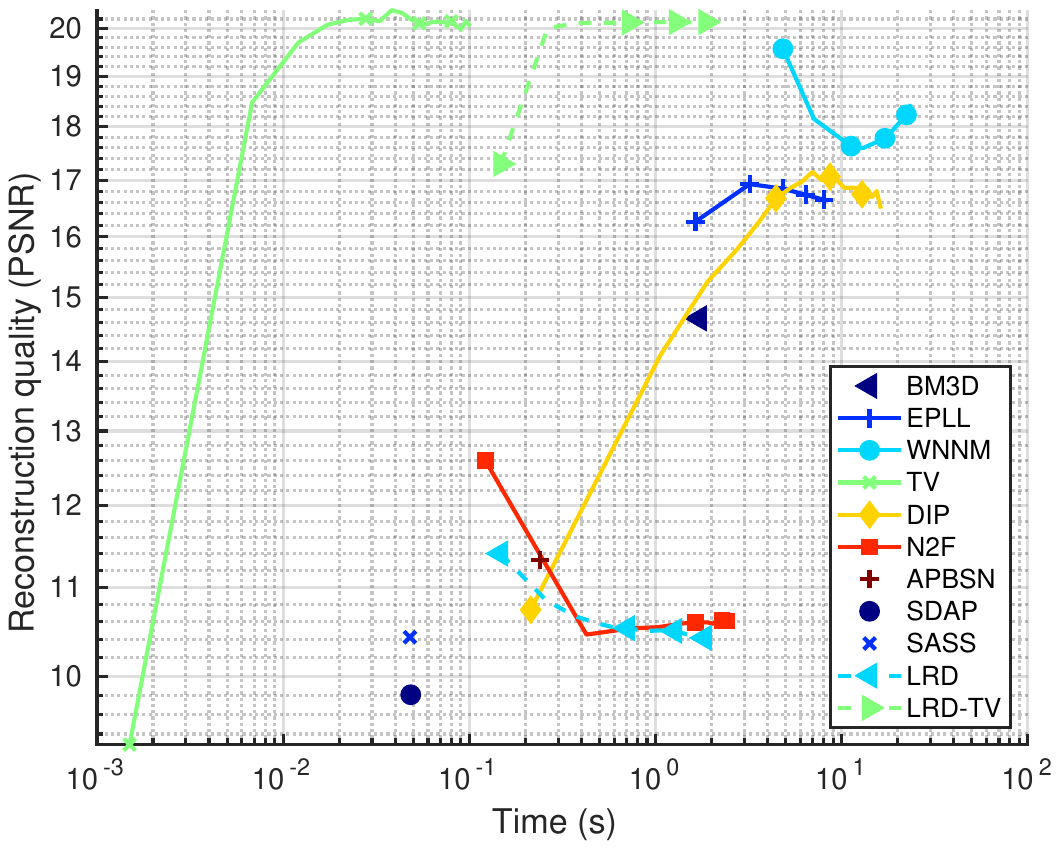}
\end{tabular}}
\vspace{-0.1cm}
\caption{\textbf{Quality of reconstruction (PSNR) vs. execution time (s).} We display overall results on the whole dataset for the four levels of input noise respectively. PSNR evolution as a function of time for a single image denoising for the chosen methods.}
\label{fig:psnr_time}
\end{figure*}

\begin{figure*}[h]
\centering
\resizebox{17.2cm}{!} {
\begin{tabular}{@{}cccccccccc@{}}
 \includegraphics[clip, angle=0]{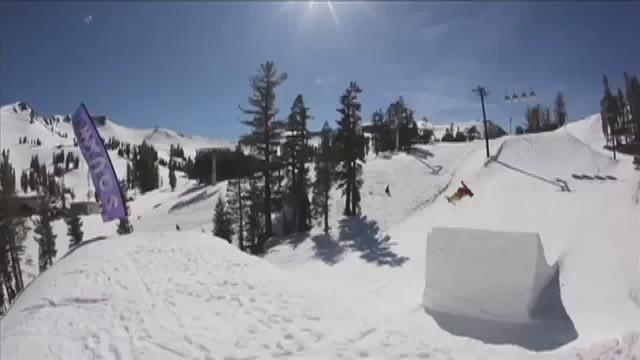}&
 \includegraphics[clip, angle=0]{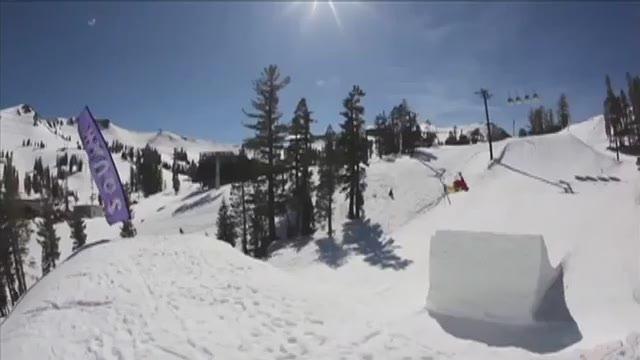}&
 \includegraphics[clip, angle=0]{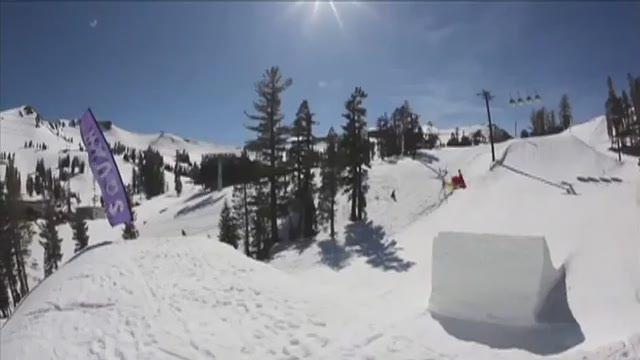}&
 \includegraphics[clip, angle=0]{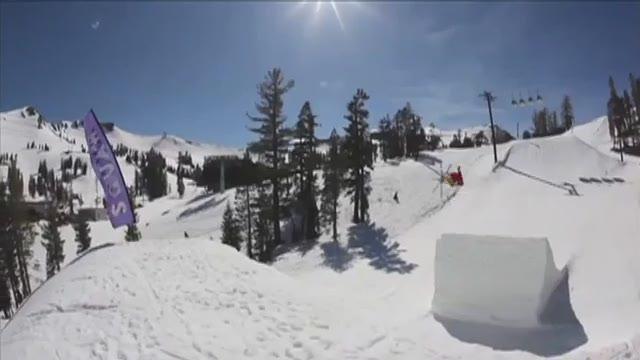}&
 \includegraphics[clip, angle=0]{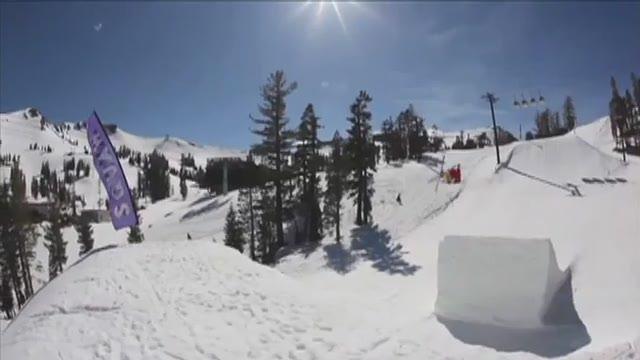}\\
  \includegraphics[clip, angle=0]{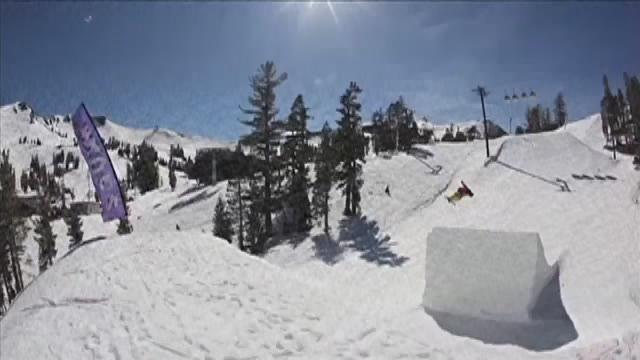}&
  \includegraphics[clip, angle=0]{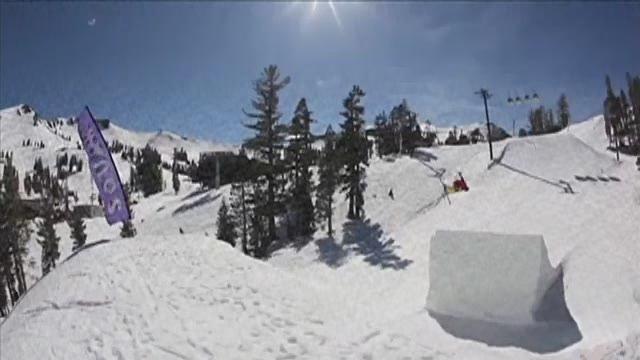}&
  \includegraphics[clip, angle=0]{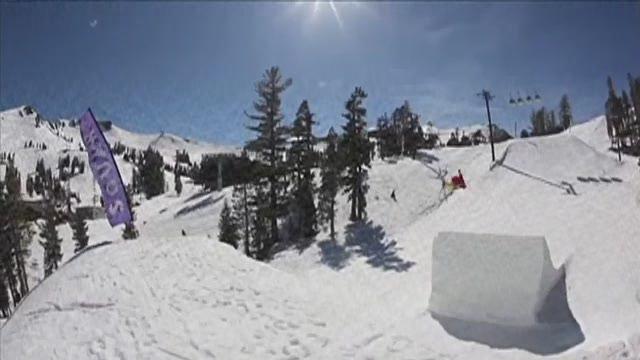}&
  \includegraphics[clip, angle=0]{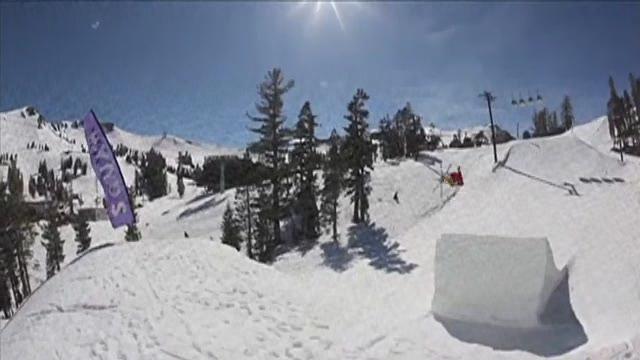}&
  \includegraphics[clip, angle=0]{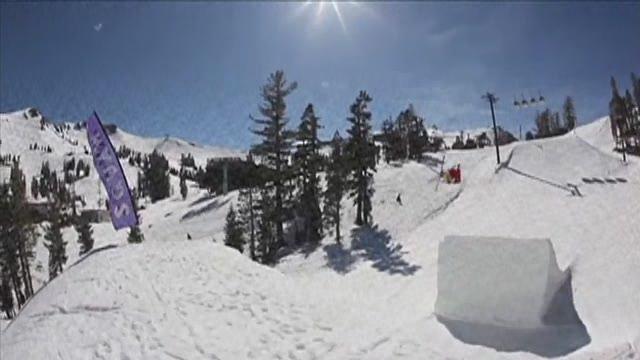}
\end{tabular}}
\vspace{-0.1cm}
  \caption{\textbf{Qualitative evaluation on RGB  video enhancement.} In all cases, we show five consecutive video frames. \textbf{Top.} Original color frames. \textbf{Bottom.} Video Enhancement. As it can be seen, our method can be used to enhance details in tensors, such video-sequences. Best viewed in color.}
\label{fig:reconstruction}
\end{figure*}

\section{Experiments}
\label{sec:exp}

\subsection{Image Denoising}

In this section, we analyse the performance of LRD-TV for image denoising problems, and we compare it to the state of the art. To that end, we select twelve gray-scale images from~\cite{liu2019image} resized to $[256\times 256]$ pixels presented in Table~\ref{table_denoising}. The methods chosen to compare to our approach are the unsupervised methods BM3D~\cite{dabov2007image}, EPLL~\cite{zoran2011learning}, TV~\cite{beck2009fast}, WNNM~\cite{gu2014weighted}, DIP~\cite{ulyanov2018deep} and LRD~\cite{reixach2023multi}, and the self-supervised N2F~\cite{lequyer2022fast}, AP-BSN~\cite{lee2022apbsn}, SDAP~\cite{Pan_2023_ICCV} and SASS~\cite{li2023spatially}. For all the methods we use authors' code and tune their parameters to achieve the best result. For the self-supervised methods, we train them with the SIDD-Small dataset for our specific noise level and use it for the evaluation together with their supplied pre-trained models, keeping the best of each for everyone of them. We consider a range of Additive White Gaussian Noise (AWGN) with $\sigma=\{30, 50, 70, 90\}$ and for each we compute the input signal quality as input PSNR (presented in the table). All experiments are carried out in a desktop computer using an INTEL\textregistered~CORE\texttrademark~i7-12700K and an NVIDIA\textregistered~GeForce\texttrademark~RTX 3090 Ti.

\begin{figure}[t!]
\centering
\resizebox{0.40\columnwidth}{!} {
 \includegraphics[viewport=50 180 560 620, clip, angle=0]{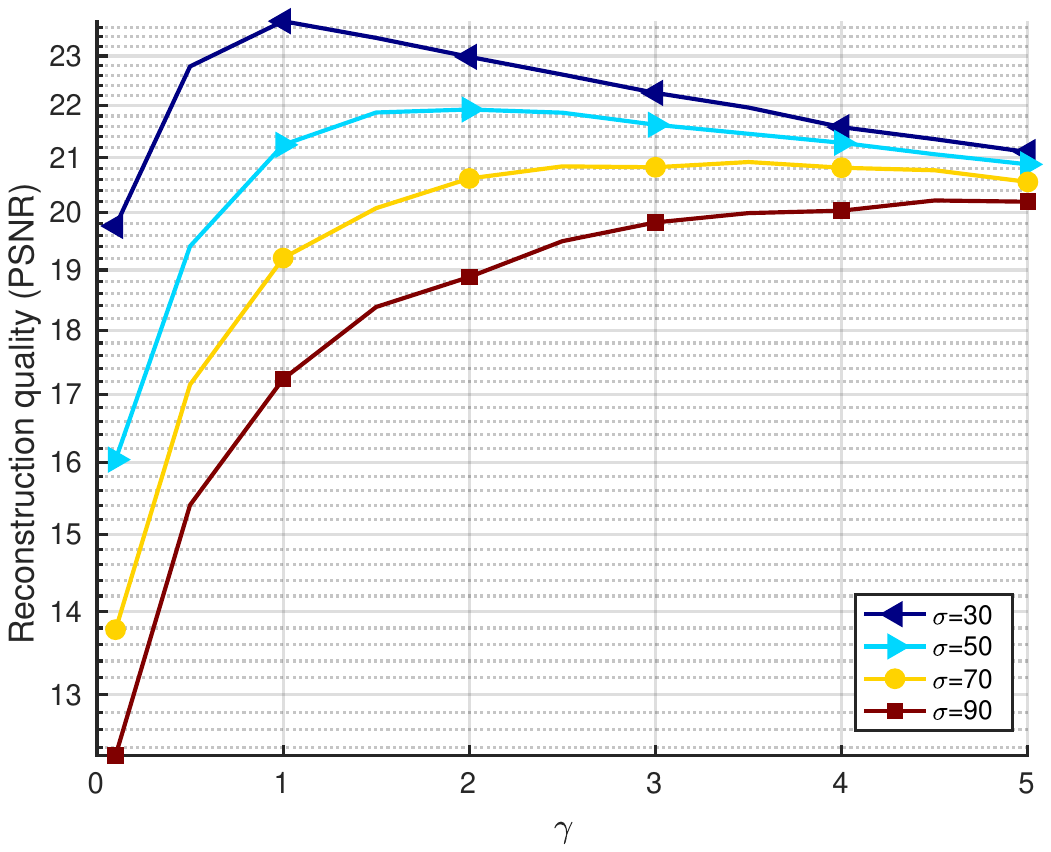}
}
\vspace{-0.1cm}
\caption{\textbf{Quality of reconstruction (PSNR) vs. $\gamma$.} We display overall results on the whole dataset for our method and the four levels of input noise.}
\label{fig:psnr_gamma}
\end{figure}

Following~\cite{reixach2023multi}, the selection of parameters for this problem is set to $\{R=3,  M=25, \alpha = 1\cdot10^{-16}\}$ 
and filters learned on the city and fruit datasets from~\cite{zeiler2010deconvolutional} with dictionary dimensions set to $\tscr{D}_m \in \mathbb{R}^{L_1\times L_2}$ for $m = \{1,\dots,M\}$ with $\{L_n = 5\}_{n=1}^2$. Regarding the choice of $\gamma$, figure~\ref{fig:psnr_gamma} reports a sensitiviy analysis from where an optimal choice can be made. For each method, we perform an image normalization after the main denoising step. The results are presented in the same table, where we have marked in blue and red
color the best and second-best achievers. We observe how LRD-TV systematically outperforms all DL approaches (DIP, N2F, AP-BSN, SDAP and SASS), performs on par with TV and is only beaten sometimes by WNNM. Our method is within the two best achievers in a majority of cases. Also, as stated in the introduction, we observe how LRD without TV regularization is not able to reject noise.

Figure~\ref{fig:psnr_time} reports a time analysis for the single image denoising task in the whole dataset for the chosen methods. As it can be seen, TV, SDAP and SASS are the most computationally efficient methods obtaining their results well below 100 ms. These are followed by LRD, LRD-TV, N2F and AP-BSN which are able to converge below 1 s. The rest of the methods require between 1 to 10 s to converge to a solution.


\subsection{Video Enhancement}

In this section we analyse the performance of LRD-TV for video enhancement problems. We select the first 10 frames of the color sequence {\em Skiing} ($[360\times 640]$ pixels) of the OTB50 dataset~\cite{wu2013online}. We represent them by means of three $3$-order tensors, one for each channel. Dictionary dimensions are chosen as $\tscr{D}_{m,c} \in \mathbb{R}^{L_1\times L_2\times L_3}$ for $m = \{1,\dots,M\}$ and $c = \{1,\dots,C\}$, with $\{L_n = 11\}_{n=1}^3 $, $M=60$ and $C=3$. We use the same data to learn the filters applying the algorithm from~\cite{garcia2018convolutional}. The rest of parameters for this problem are set to $\{R=16, \alpha = 1\cdot10^{-16}, \gamma = 1\cdot10^{-3}, \zeta = 5\cdot10^{-3}\}$, and $\{\gamma_m = 0\}_{m=1}^{30}$, $\{\gamma_m = \gamma\}_{m=31}^{60}$ and $\{\zeta_m = \zeta\}_{m=1}^{30}$, $\{\zeta_m = 0\}_{m=31}^{60}$. The reconstruction parameters are: $\delta = 0.6$, $\{\delta_m = \delta\}_{m=1}^{30}$, $\{\delta_m = 0\}_{m=31}^{60}$.

Figure~\ref{fig:reconstruction} presents the qualitative results for the first five frames of the sequence. We can observe how in the ski slope snow details appear improved and in the sky, clouds texture is remarked, resulting in a visually more complex scene.


\section{Conclusion}

LRD is a powerful framework that eases the inclusion of priors to its formulation making it able to solve tensor restoration problems. The results regarding image denoising and video enhancement verify our claims and prove that TV regularization can still compete against the state of the art. The method takes advantage of being an analytical approach and does not require an extensive training stage like most DL approaches.

\pagebreak

\bibliographystyle{IEEEbib}
\bibliography{refs}

\end{document}

%% file: newcommands2.tex
\newcommand{\comment}[1]{}

\newcommand{\bs}{\mathbf{s}}

\newcommand{\bt}{\mathbf{t}}
\newcommand{\bu}{\mathbf{u}}

\newcommand{\bv}{\mathbf{v}}

\newcommand{\bx}{\mathbf{x}}

\newcommand{\bA}{\mathbf{A}}

\newcommand{\bD}{\mathbf{D}}

\newcommand{\bI}{\mathbf{I}}

\newcommand{\bK}{\mathbf{K}}

\newcommand{\bQ}{\mathbf{Q}}

\newcommand{\bS}{\mathbf{S}}

\newcommand{\bW}{\mathbf{W}}
\newcommand{\bX}{\mathbf{X}}

\newcommand{\argmin}{\operatornamewithlimits{arg\,min}}

%% file: newcommands3.tex
\makeatletter
\DeclareRobustCommand\onedot{\futurelet\@let@token\@onedot}
\def\@onedot{\ifx\@let@token.\else.\null\fi\xspace}

\def\ie{\emph{i.e}\onedot}

\makeatother


%% file: paper.bbl
\begin{thebibliography}{10}

\bibitem{rudin1992nonlinear}
Leonid~I Rudin, Stanley Osher, and Emad Fatemi,
\newblock ``Nonlinear total variation based noise removal algorithms,''
\newblock {\em Physica D: nonlinear phenomena}, vol. 60, no. 1-4, pp. 259--268, 1992.

\bibitem{elad2006image}
M.~Elad and M.~Aharon,
\newblock ``Image denoising via sparse and redundant representations over learned dictionaries,''
\newblock {\em TIP}, vol. 15, no. 12, pp. 3736--3745, 2006.

\bibitem{dabov2007image}
Kostadin Dabov, Alessandro Foi, Vladimir Katkovnik, and Karen Egiazarian,
\newblock ``Image denoising by sparse 3-d transform-domain collaborative filtering,''
\newblock {\em IEEE Transactions on image processing}, vol. 16, no. 8, pp. 2080--2095, 2007.

\bibitem{heide2015fast}
F.~Heide, W.~Heidrich, and G.~Wetzstein,
\newblock ``Fast and flexible convolutional sparse coding,''
\newblock in {\em Proceedings of the IEEE Conference on Computer Vision and Pattern Recognition}, 2015, pp. 5135--5143.

\bibitem{papyan2017convolutional}
V.~Papyan, Y.~Romano, J.~Sulam, and M.~Elad,
\newblock ``Convolutional dictionary learning via local processing,''
\newblock in {\em ICCV}, 2017, pp. 5296--5304.

\bibitem{ulyanov2018deep}
Dmitry Ulyanov, Andrea Vedaldi, and Victor Lempitsky,
\newblock ``Deep image prior,''
\newblock in {\em Proceedings of the IEEE conference on computer vision and pattern recognition}, 2018, pp. 9446--9454.

\bibitem{lequyer2022fast}
Jason Lequyer, Reuben Philip, Amit Sharma, Wen-Hsin Hsu, and Laurence Pelletier,
\newblock ``A fast blind zero-shot denoiser,''
\newblock {\em Nature Machine Intelligence}, vol. 4, no. 11, pp. 953--963, 2022.

\bibitem{lee2022apbsn}
Wooseok Lee, Sanghyun Son, and Kyoung~Mu Lee,
\newblock ``Ap-bsn: Self-supervised denoising for real-world images via asymmetric pd and blind-spot network,''
\newblock in {\em Proceedings of the IEEE/CVF Conference on Computer Vision and Pattern Recognition (CVPR)}, 2022.

\bibitem{li2023spatially}
Junyi Li, Zhilu Zhang, Xiaoyu Liu, Chaoyu Feng, Xiaotao Wang, Lei Lei, and Wangmeng Zuo,
\newblock ``Spatially adaptive self-supervised learning for real-world image denoising,''
\newblock in {\em Proceedings of the IEEE/CVF Conference on Computer Vision and Pattern Recognition}, 2023.

\bibitem{Pan_2023_ICCV}
Yizhong Pan, Xiao Liu, Xiangyu Liao, Yuanzhouhan Cao, and Chao Ren,
\newblock ``Random sub-samples generation for self-supervised real image denoising,''
\newblock in {\em Proceedings of the IEEE/CVF International Conference on Computer Vision (ICCV)}, October 2023, pp. 12150--12159.

\bibitem{reixach2023multi}
David Reixach,
\newblock ``Multi-dimensional signal recovery using low-rank deconvolution,''
\newblock in {\em ICASSP 2023-2023 IEEE International Conference on Acoustics, Speech and Signal Processing (ICASSP)}. IEEE, 2023, pp. 1--5.

\bibitem{beck2009fast}
Amir Beck and Marc Teboulle,
\newblock ``Fast gradient-based algorithms for constrained total variation image denoising and deblurring problems,''
\newblock {\em IEEE transactions on image processing}, vol. 18, no. 11, pp. 2419--2434, 2009.

\bibitem{liu2019image}
Jiaming Liu, Yu~Sun, Xiaojian Xu, and Ulugbek~S Kamilov,
\newblock ``Image restoration using total variation regularized deep image prior,''
\newblock in {\em ICASSP 2019-2019 IEEE International Conference on Acoustics, Speech and Signal Processing (ICASSP)}. Ieee, 2019, pp. 7715--7719.

\bibitem{zoran2011learning}
Daniel Zoran and Yair Weiss,
\newblock ``From learning models of natural image patches to whole image restoration,''
\newblock in {\em 2011 international conference on computer vision}. IEEE, 2011, pp. 479--486.

\bibitem{gu2014weighted}
Shuhang Gu, Lei Zhang, Wangmeng Zuo, and Xiangchu Feng,
\newblock ``Weighted nuclear norm minimization with application to image denoising,''
\newblock in {\em Proceedings of the IEEE conference on computer vision and pattern recognition}, 2014, pp. 2862--2869.

\bibitem{zhang2017learning}
Kai Zhang, Wangmeng Zuo, Shuhang Gu, and Lei Zhang,
\newblock ``Learning deep cnn denoiser prior for image restoration,''
\newblock in {\em Proceedings of the IEEE conference on computer vision and pattern recognition}, 2017, pp. 3929--3938.

\bibitem{harshman1970foundations}
R.~A. Harshman et~al.,
\newblock ``Foundations of the {PARAFAC} procedure: Models and conditions for an "explanatory" multimodal factor analysis,''
\newblock 1970.

\bibitem{carroll1970analysis}
J.~D. Carroll and J.~J. Chang,
\newblock ``Analysis of individual differences in multidimensional scaling via an n-way generalization of “eckart-young” decomposition,''
\newblock {\em Psychometrika}, vol. 35, no. 3, pp. 283--319, 1970.

\bibitem{kolda2006multilinear}
T.~G. Kolda,
\newblock ``Multilinear operators for higher-order decompositions.,''
\newblock Tech. {R}ep., Sandia National Laboratories, 2006.

\bibitem{bader2006algorithm}
B.~W. Bader and T.~G. Kolda,
\newblock ``Algorithm 862: {MATLAB} tensor classes for fast algorithm prototyping,''
\newblock {\em TOMS}, vol. 32, no. 4, pp. 635--653, 2006.

\bibitem{zeiler2010deconvolutional}
M.~D. Zeiler, D.~Krishnan, G.~W. Taylor, and R.~Fergus,
\newblock ``Deconvolutional networks,''
\newblock in {\em CVPR}, 2010, pp. 2528--2535.

\bibitem{wu2013online}
Y.~Wu, J.~Lim, and M.~H. Yang,
\newblock ``Online object tracking: A benchmark,''
\newblock in {\em CVPR}, 2013, pp. 2411--2418.

\bibitem{garcia2018convolutional}
C.~Garcia-Cardona and B.~Wohlberg,
\newblock ``Convolutional dictionary learning: A comparative review and new algorithms,''
\newblock {\em TCI}, vol. 4, no. 3, pp. 366--381, 2018.

\end{thebibliography}
